\documentclass[twoside,twocolumn]{article}

\usepackage[sc]{mathpazo} 
\usepackage[T1]{fontenc} 
\usepackage{microtype} 

\usepackage[english]{babel} 

\usepackage[hmarginratio=1:1,top=32mm,columnsep=20pt]{geometry} 
\usepackage[hang, small,labelfont=bf,up,textfont=it,up]{caption} 
\usepackage{booktabs} 

\usepackage{lettrine} 

\usepackage{enumitem} 
\setlist[itemize]{noitemsep} 

\usepackage{abstract} 

\usepackage{titlesec} 

\titleformat{\section}[block]{\large\scshape\centering}{\thesection.}{1em}{} 
\titleformat{\subsection}[block]{\large}{\thesubsection.}{1em}{} 

\usepackage{fancyhdr} 
\usepackage{titling} 

\usepackage{hyperref} 
\usepackage{graphicx} 

\pretitle{\begin{center}\Huge\bfseries} 
\posttitle{\end{center}} 
\title{Active learning to optimise time-expensive algorithm selection} 
\author{%
\textsc{Riccardo Volpato}\thanks{Corresponding author} \\[1ex] 
\normalsize Satalia \\ 
\normalsize \href{mailto:riccardo@satalia.com}{riccardo@satalia.com} 
\and 
\textsc{Guangyan Song}\\[1ex] 
\normalsize Satalia \\ 
\normalsize \href{mailto:guangyan@satalia.com}{guangyan@satalia.com} 
}
\date{August 2017} 
\renewcommand{\maketitlehookd}{%
\begin{abstract}
\noindent Hard optimisation problems such as Boolean Satisfiability typically have long solving times and can usually be solved by many algorithms, although the performance can vary widely in practice. Research has shown that no single algorithm outperforms all the others; thus, it is crucial to select the best algorithm for a given problem. Supervised machine learning models can accurately predict which solver is best for a given problem, but they require first to run every solver in the portfolio for all examples available to create labelled data. As this approach cannot scale, we developed an active learning framework that addresses this problem by constructing an optimal training set, so that the learner can achieve higher or equal performances with less training data. Our work proves that active learning is beneficial for algorithm selection techniques and provides practical guidance to incorporate into existing systems.
\end{abstract}
}

\begin{document}

\maketitle


\section{Introduction}
The algorithm selection problem is defined as the automated selection of the algorithm that best suits a problem given a metric to optimise \cite{Rice:76}. Algorithm selection traditionally works by extracting specific features from a new problem and learning over time the relationship between these features and which algorithm is best for the problem \cite{Gomes:01}. However, with optimisation algorithms (also known as solvers), this is a costly learning process because to identify the fastest solver demands to run every solver available. Only problems that are compared across all solvers can be used to train the algorithm selector. This study focuses on a specific optimisation problem known as Boolean Satisfiability problems (SAT). Building on top of the existing research \cite{Xu:12a} and available SAT solver data, it concentrates on the development of algorithm selection for SAT problems. Research has shown that supervised machine learning models can accurately predict which solver is best for a given SAT problem \cite{Xu:07} but they require collecting a set of problems each one associated with a solver that solves it in the shortest time possible (the best solver). Finding a problem's best solver requires to run each solver on the specific problem and return the solver associated with the shortest solving time. Clearly, the more solvers considered, the longer it takes to label a single instance. Additionally, SAT problems are NP-hard, which means that, in practice, solving a problem with many variables can take up to several hours of computation. This makes labelling SAT problems very expensive, whereas parsing the problem and computing its features is usually computationally cheap. The computation complexity of solving SAT problems lead to the following challenges in predicting a problem's best solvers:
\begin{itemize}
\item Gathering training data is very time consuming and often un-feasible if lacking a cluster of powerful machines (as used in the SAT competitions). This presents a challenge for the training of data-greedy methods which usually performs better with large training samples.
\item Since labelling time is a scarce resource, selecting which unlabelled problem to label and integrate into the training set becomes a strategic decision that can be optimised.
\end{itemize}

Our contribution consists of developing a novel active learning framework to optimise the generation of time-expensive algorithm selection training sets. Active learning is a machine learning methodology commonly employed in the presence of abundant problem samples that are hard to label. Active learning models provide a clever guideline to build the optimal training set of labelled instances, by selecting which among the many unlabelled ones are best to query. We validate our approach using data from SAT competitions and show that we can build better or equally good algorithm selectors using less training data and saving precious solving time.

The remainder of this article is organised as follows. Section~\ref{related work} gives an account of previous work. Our active learning framework is described in Section~\ref{approach} and experimental results are presented in Section~\ref{results}. Finally, Section~\ref{conclusions} gives the conclusions.


\section{Related work}\label{related work}
Optimisation focuses on finding the most efficient method to solve a problem. However often there is not a single algorithm that can outperform all others in solving every problem. Research has shown that different instances of the same classes are solved faster by different solvers \cite{Gomes:01}. The theory refers to a concept called virtual best solver (VBS) which solves optimally a set of problems by running the best existing algorithm for each problem. Unfortunately, this is just a theoretical oracle that does not exist. In practice, identifying the best solver for a problem mandates to know beforehand the run-time of each solver. Hence, the strategy of most existing approaches is using heuristics to approximate the VSB.

IIn the context of Boolean Satisfiability problems, the most widely adopted methodology was the winner-take-all approach. This involves running all the solvers on a set of training problems and then, for any future problem, use the solver that performed best on average in the training set. However, the context of SAT competitions proved the existence of many solvers that despite having poor performances on average offer very good on particular instances. Because the winner-take-all approach cannot leverage these instance-specific solvers, it struggles to approximate well the VBS \cite{Xu:07}. During the last decade, SATzilla \cite{Nudelman:04} pioneered a new machine learning-based methodology that showed significant improvements over all the previous approaches. Given a set of problem features, the algorithm portfolio method selects the best solver for the given problem. In this fashion, the algorithm can use different solvers for different problems, mimicking better the idea of the VBS. Although extracting problem features requires extra time, this cost has proven to be empirically worth in term of overall solving time \cite{Xu:07}. SATzilla won five medals in both the 2007 and 2009 SAT competitions and won the 2012 SAT challenge \cite{Balint:12}. After establishing SATzilla's effectiveness, the authors decided to not compete in the main track of the 2011 SAT competition, to avoid daunting new work on non-portfolio solvers \cite{Xu:12a}. Since 2004, SATzilla methodology evolved widely. The authors employed both of the two main ways to construct an algorithm selection portfolio: regression of run-times and classification of the best solver. A regression algorithm portfolio works by predicting the run- time of each solver in the portfolio and selecting the solver having the smallest predicted run-time. This requires to build $n$ regression models, one for each of the $n$ solvers in the portfolio. Consequently, regression models provide a precise estimate of each solver running time. Besides identifying the best solver, this information can be used to rank the solvers by expected performance and use the second-best solver in case the first one fails. The original SATzilla-07 was a portfolio of empirical hardness models, computationally inexpensive predictors of an algorithm's run-time based on features of a given problem and the algorithm's past performance \cite{Leyton-Brown:02}. Each hardness model predicted a solver's run-time using a Ridge Regression over the problem features conditional upon if the given problem was satisfiable or not, which was conversely predicted using a Sparse Multinomial Logistic Regression (SMLR) \cite{Krishnapuram:05}. It must be mentioned that when a solver keeps running indefinitely without returning a solution, it is usually stopped at a certain cutoff time. This data, which is not the real solving time of the solver, is right-censored and can bias the regression. Censored data are heavily studied in Survival analysis which provides numerous tools to tame their effects. SATzilla-07 proved that estimating censored data using Schmee and Hahn \cite{Schmee:79} method performs better than deleting them or using the cutoff time as an estimate of the actual solving time. Conversely, building a classification algorithm portfolio overcomes the issue of right-censored data. A classification portfolio simply predicts which solver is best for solving a given problem. There are several ways to build a classification portfolio. The most straight forward approach is to compare all run-times for a given instance and label it with the name of the best performing solver. For instance, if a problem i is solved in 420 seconds by $solver_1$, 599 seconds by $solver_2$ and 187 seconds by $solver_3$, the problem is labelled as $solver_3$. A possible alternative \cite{Matos:14} is to build a binary classifier that predicts if a solver is "bad" (0) or "good" (1) for a given problem. The goodness of the solver can be established using different criteria such as whether the solvers is faster than the feature computation algorithm or if the solver is faster than the solver identified by the winner-takes-all approach. A third method is to build $n$ pairwise classifiers for the $n$ solvers in the portfolio and to predict the best solver in each pair. The best solver is the one that receives the majority of votes across all pairwise classifiers. SATzilla 2012 \cite{Xu:12a} implemented the third method using cost-sensitive pairwise decision forests, punishing errors in classification in direct proportion to their impact on the performances of the portfolio. Both regression and classification versions of SATzilla include pre-solvers and back-up solvers. Pre-solvers are automatically determined on a training set of problems and later run for a short amount of time before features are computed. They ensure good performances on very easy instances and allow the predictive models to focus exclusively on harder instances. The back-up solver, identified using a validation set, is used for all instances for which pre-solvers and features computation time out. Differently, from most of the work mentioned above, the purpose of this study is to explore machine learning methods beyond the supervised approaches. Given the exploratory nature of this study, we did not include pre-solvers and back-up solvers in our experimentation.

\section{Approach}\label{approach}
To extract features necessary to learn the best solver for a given problem, we used the approached developed within \cite{Xu:12b}. Features 1-7 concern the numbers of clauses and variables and how formula simplification using the SATelite solver reduces them. Features 8-36 are graph features. Features 37-46 regard the ratio of positive to negative literal per variable and clause. Features 47-49 count the number of unary, binary and ternary clauses and features 50-55 are statistics regarding Horn clauses. Features 56-62 estimate the hardness of the search space for a Davis-Putnam-Logemann-Loveland (DPLL) solver. Measured statistics are the number of unit propagation at various depth for a DPLL procedure and the size of the search space generated by randomly selecting variables and doing unit propagation until a contradiction emerges. Features 63-66 are obtained by solving a linear relaxation of an integer program representing the SAT problem. Features 69-90 are based on two local search solvers: GSAT and SAPS \cite{Tompkins:05}. Features 91-108 measures the number and length of clause learned by running the Zchaff-rand solver \cite{Mahajan:05} for 2 seconds. Features 109-126 compute statistics regarding estimated probabilities of each variable required to be true, false or unconstrained \cite{Hsu:08}. Finally, features 128-138 measure the time required for the computation of 12 different blocks of codes used to compute the features described above.

After feature extraction, a matrix of problems with the features above as columns undergoes the following pre-processing steps:
\begin{enumerate}
\item  Remove uninformative columns (i.e. standard deviation of the feature $\sigma = 0$)
\item Remove uninformative problems that have an average solving time lower than 0.01 seconds or are not solved by any solver in the portfolio
\item Standardise the feature values and imputing missing values using a k-nearest-neighbours (k-NN) alike procedure (see Appendix~\ref{appendix-A})
\end{enumerate}

Finally, our proposed approach outlines a way to build the best possible training set $T$ in a large pool of $P$ of unlabelled SAT problems. To use knowledge of $T$ over $P$ the approaches uses as learner $\mathcal{L}$ an active learning classification random forest. Both regression and classification random forest were considered as an option for the proposed approach. However, since the active learning framework that employed requires to predict class probabilities for each unlabelled problem, the final methodology uses a classification random forest. This choice of classification over regression also permits to safely ignore the censored data problem and was empirically validated by the latest works in SAT portfolio solvers \cite{Xu:12a}. The random forest algorithm \cite{Breiman:01} works by building a randomised ensemble of decision trees. Decision trees are composed by splitting nodes and leaves. A splitting node uses a specific value of a specific attribute to partition the data, whereas a leaf node identifies an output class. We construct our random forest by recursively splitting the data on the attribute $A$ that maximise the entropy gain criteria \cite{Shannon:48}. To construct the random forest, we build each tree using the above procedure on a random sample of features, which we set as $|log(k) + 1|$ where $k$ is the number of feature in the data-set (approximately 138 in our experiments). We set the forest size as 99, which is usually recommended in the relevant literature \cite{Xu:12a}. For a given instance, using the values of its attributes we can follow the tree until we end in leaf, which returns the class predicted for the given instance. The predicted class is the one with the highest mean probability estimate across all trees. We can also compute the predicted class probabilities of the sample as the mean predicted class probabilities across all trees.

Active learning (also known as query learning) was developed on the concept that if a learning algorithm is allowed to choose the data from which it will learn, it will perform better with less training \cite{Settles:09}. Allowing the learner to choose its learning data first requires a sampling method. Different methods, such as membership query synthesis or pool-based sampling, suit different scenarios \cite{Settles:09}. For this study, we employed selective sampling \cite{Cohn:94}. Selective sampling is based on the assumption that the learner can access all the unlabelled data and decide whether or not to query it for labelling. In our context, the random forest can predict probabilities over all the unlabelled SAT problems and given the predictions over the select the problems to query for its learning. Given a sampling method, there are different query strategy framework that can be used to select which instances to label and which not. Our framework uses the minimum margin uncertainty sampling. Uncertainty sampling \cite{Lewis:94} consists of querying the instance on which the learner is most uncertain about and is often straightforward for probabilistic methods. For learner $\theta$, features $x$ and labels $y$ the minimum margin sampling criteria \cite{Scheffer:01} measure uncertainty in the following way:

$$
x_{MS}^* = \arg\min_{x \in \mathcal{U}} [ P_{\theta} (\hat{y}_1 | x) - P_{\theta} (\hat{y}_2 | x)]
$$

Other sampling criteria to measure uncertainty are maximum uncertainty:

$$
x_{MU}^* = \arg\max_{x \in \mathcal{U}} [ 1 - \arg\max_y P_{\theta} (\hat{y} | x)]
$$

and maximum entropy \cite{Shannon:48}:

$$
x_{H}^* = \arg\max_{x \in \mathcal{U}} [ -\sum_i  P_{\theta} (\hat{y_i} | x) \log{P_{\theta} (\hat{y_i} | x)}]
$$

We compare different active learning schemes in our experiment in Section~\ref{results} and show that margin sampling is the most effective choice is the majority of settings.

\begin{figure*}[ht]
  \centering
  \includegraphics[width=0.8\linewidth]{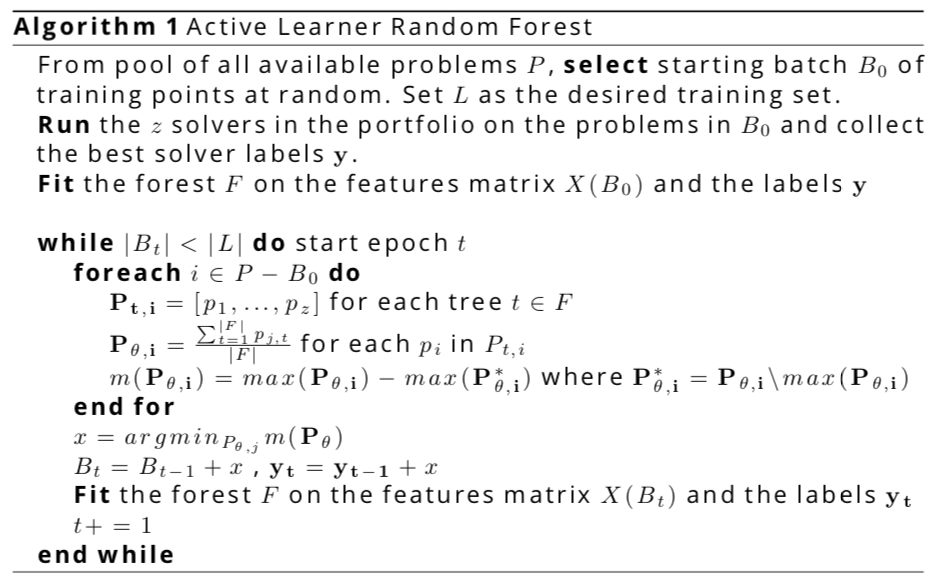}
  \caption{Active learning random forest}
  \label{fig:algorithm1}
\end{figure*}

Algorithm \ref{fig:algorithm1} summarises our proposed approach. To achieve good performance, the size of the starting batch $B_0$ usually equals $\frac{|L|}{3}$ or $\frac{|L|}{10}$ , where $L$ the full training set that will be used for the classification. However, this value can be optimised as a trade-off between desired performances and cost of training the starting batch.

\section{Results}\label{results}
The approach proposed in section~\ref{approach} was tested against alternative approaches on a dataset of SAT problems. After we verified that a multi-class random forest is a sensible method for selecting SAT solvers by benchmarking its predictive performances against other methods, we tested if the suggested active learning adapted random forest showed significant gains over a traditional passive learner.

\subsection{Experimental Setup}\label{res-setup}
The data used for the experiments of this study is a publicly available industrial SAT instances dataset used in the evaluation of SATzilla 2012. It is based on a collection of industrial SAT instances from all three SAT competitions and three SAT Races between 2006 and 2012. It displays runtimes for 31 SAT solvers - with a cutoff time of 1200 seconds. Instances that could not be solved by any of the 31 solvers were dropped, leaving a total of 1167 instances.

Among these 31 solvers, we performed experiments used different numbers of solvers to verify that the results were not due mainly to the size of solvers in the portfolio. Specifically, we performed experiments using a solver portfolio of size 3 (using \textit{Glucose}, \textit{Minisat} and \textit{Lingeling} solvers), size 6 (the previous 3 plus \textit{Restartsat}, \textit{Lrgshr} and \textit{Mxc09}) and size 10 (the previous 6 plus \textit{Rcl}, \textit{Precosat}, \textit{MphaseSAT64} and \textit{Qutersat}). The specific solvers were those that had the best performances on average across all the problems in the dataset.
 
When slicing the data for specific sets of solvers we excluded problems whose runtime across solvers had a mean and standard deviation of runtime lower than 0.01 seconds. For instance, when predicting the best solver among \textit{Glucose}, \textit{Minisat} and \textit{Lingeling} this procedure reduced the dataset to 897 samples. Additionally, we set the k-NN inputter method (see Appendix~\ref{appendix-A}) to 3 nearest neighbours. This resulted in the imputation of 13\% feature values.

To validate the performance of the recommended approach, we compared the random forest classifier of 99  trees built by randomly selecting $log(k) + 1$ out of the $k$ features available against two other methods that returned the best performances among a larger pool of tested methods. These are:
\begin{itemize}
\item{A multi-class logistic regression with a cross-entropy loss function using the Newton-CG optimisation algorithm and a regularisation strength parameter of 1/100.}
\item{A multi-layer perceptron combined with two recurrent layers referred to as RNN.}
\end{itemize}
 
Performances were assessed using the following metrics:
\begin{itemize}
\item{\textit{Simple accuracy (acc)}: number of times learner correctly identify the best solver out of the number of total attempts}
\item{\textit{Mean size of errors (mes)}:  the average of the absolute difference between predicted solver runtime and virtual best solver runtime as a time value and as a percentage of the virtual best solver (VBS) average runtime. The VBS runtime is the sum of the runtimes of each best solver for each problem. The average is obtained dividing this sum by the number of problems.}
\item{\textit{Lenient accuracy 5 seconds (acc5)}: number of times learner select a solver that solved the problem within 5 seconds of the virtual best solver.}
\end{itemize}

\subsection{Passive Learners}\label{res-passive}
In order to have more confident results test on overall methods used 10-fold cross validation. Table 1 reports the accuracy of predicting the best solver among 3 solvers. Table 2 extend the same evaluation to three more solvers, and table 3 to four more, for a total of 10 solvers. Results from predictions over 3 solvers confirm that random forest is the best performing approach. Lenient accuracy results are very encouraging: the selected solver solves the problem within 5 seconds of the VBS solving time in over 85\% of the cases.

\begin{table}
\caption{Best solver prediction performances for 3 solvers (VBS average runtime of 110.8 seconds)}
\centering
\begin{tabular}{lrrrr}
\toprule
\cmidrule(r){1-2}
Model & acc & mes (sec) &  mes\% & acc5\\
\midrule
Logit & 61.9\% & 43.9 & 39.6\% & 77.6\% \\
RNN & 70.3\% & 33.4 & 30.1\% & 83.5\% \\
RF & 73.4\% & 18.8 & 17\% & 87\% \\
\bottomrule
\end{tabular}
\end{table}

\begin{table}
\caption{Best solver prediction performances for 6 solvers (VBS average runtime of 103.9 seconds)}
\centering
\begin{tabular}{lrrrr}
\toprule
\cmidrule(r){1-2}
Model & acc & mes (sec) &  mes\% & acc5\\
\midrule
Logit & 45\% & 76.5 & 73.6\% & 66\% \\
RNN & 48\% & 67.4 & 64.8\% & 69.1\% \\
RF & 64.3\% & 31.7 & 30.5\% & 82.4\% \\
\bottomrule
\end{tabular}
\end{table}

\begin{table}
\caption{Best solver prediction performances for 10 solvers (VBS average runtime of 96.9 seconds)}
\centering
\begin{tabular}{lrrrr}
\toprule
\cmidrule(r){1-2}
Model & acc & mes (sec) &  mes\% & acc5\\
\midrule
Logit & 43.1\% & 98.9 & 102\% & 63.3\% \\
RNN & 40.7\% & 81.2 & 83.7\% & 62.5\% \\
RF & 63.1\% & 35.2 & 36.4\% & 81.6\% \\
\bottomrule
\end{tabular}
\end{table}

As expected, increasing the number of solvers to choose from generated more noise and the performances of all algorithms dropped. However, the random forest is still the best performing method and the drop in the lenient accuracy is half the drop in the overall accuracy, suggesting that the approach is often able to make sub-optimal but still acceptable (in terms of runtime) predictions. Especially for the random forest classifier, expanding the solver portfolio from 6 to 10 solvers has almost no effect on the performances, which shows that the approach is robust for larger solver portfolios. The average VBS decreases while increasing the size of the portfolio solver because instances for which a newly inserted solver is the best solver have a lower VBS runtime in the experiment with more solvers. Additionally, instances that are not solved by any solvers in the smaller portfolio but are solved by one of the newly added solvers are added to the dataset, usually increasing its size. For this reason, we also included the absolute value of the mean error size \textit{(mes)} as the percentage \textit{mes} is not comparable across experiments.

\subsection{Active Learners}\label{res-active}
\begin{figure*}[ht]
  \centering
  \includegraphics[width=0.8\linewidth]{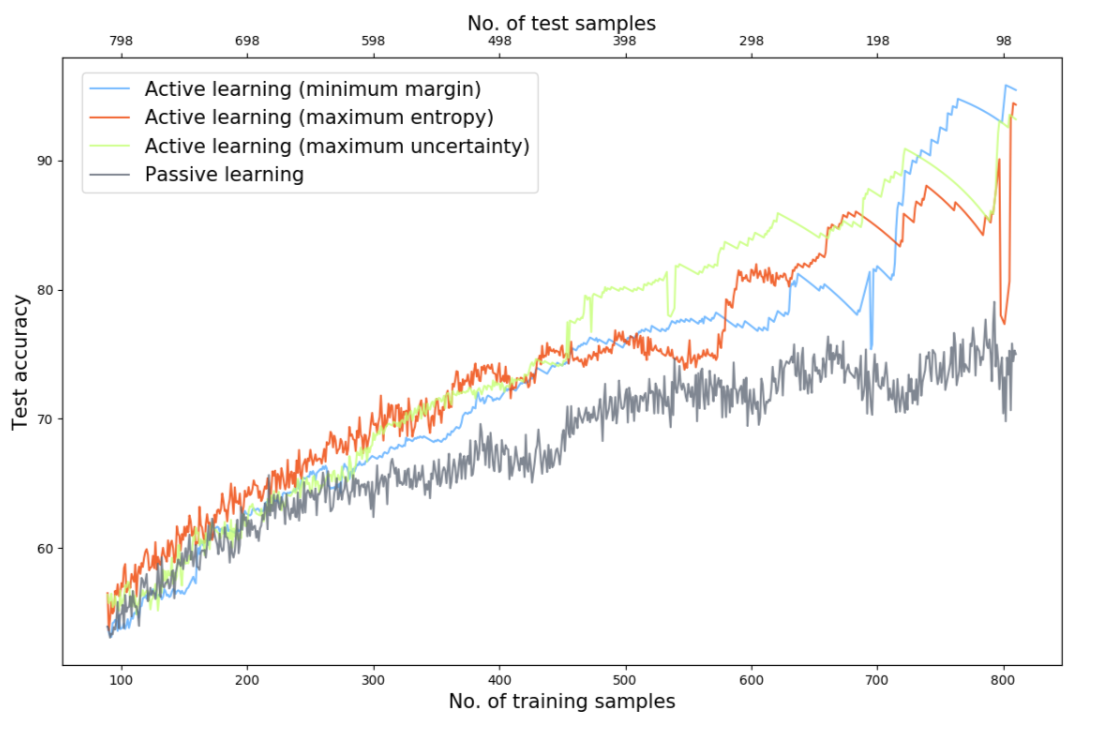}
  \caption{Test accuracy for active and passive learners with 3 SAT solvers}
  \label{fig:figure-acc3}
\end{figure*}

\begin{figure*}[ht]
  \centering
  \includegraphics[width=0.8\linewidth]{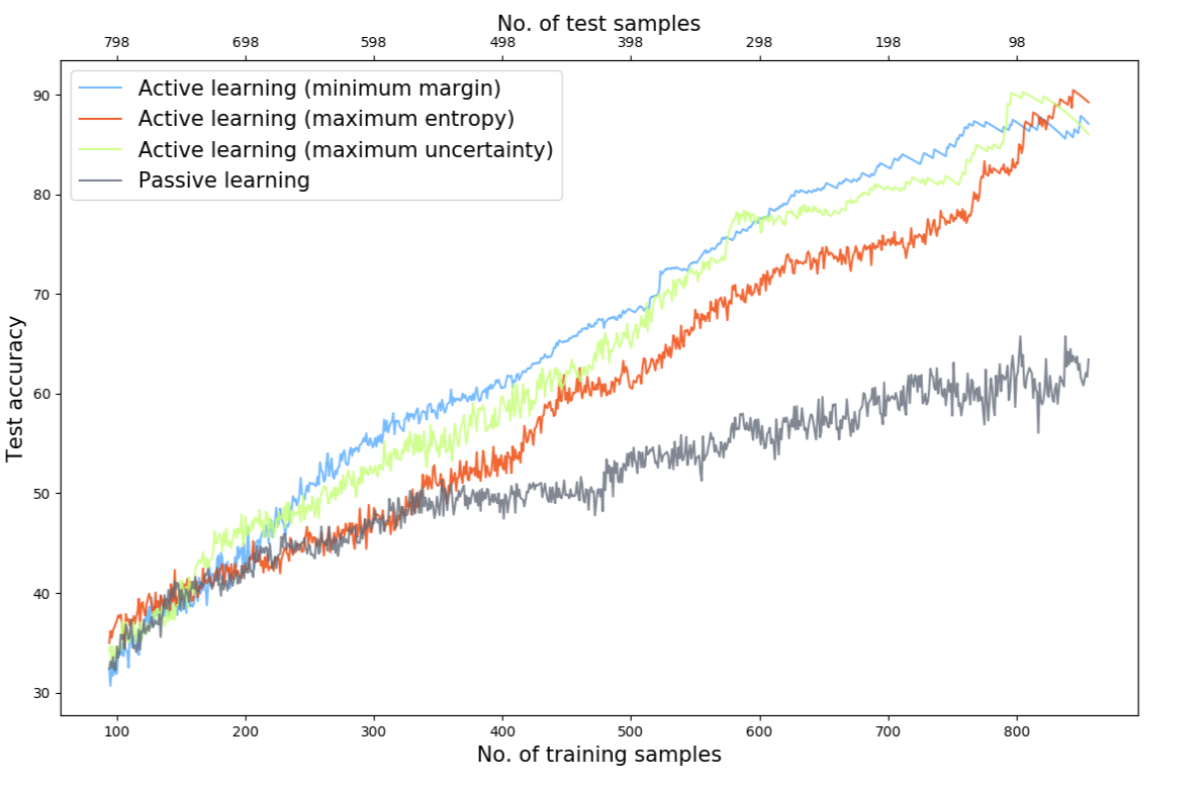}
  \caption{Test accuracy for active and passive learners with 10 SAT solvers}
  \label{fig:figure-acc10}
\end{figure*}

As described above, experiments on overall methods confirmed the intuition of \cite{Xu:12a} that random forest outperforms most of the other methods in identifying the best solver. Based on this, we experimented modifying the learning process of the forest using several active learning schemes. Following the logic of the proposed method in Algorithm \ref{fig:algorithm1}, after sampling a tenth of the dataset as $B0$ to train the learner, we iteratively increased the training set using both the Active Learner Random Forest as well as random sampling Random Forest (passive learner).

Figure \ref{fig:figure-acc3} shows the test accuracy for training-test sets of different sizes constructed using both active and passive learning. The line charts show that active learning has a clear advantage over passive learning. It is evident that for most of the training-test set sizes the active learning random forest achieves higher predictive performances. In a specular manner, active learning can achieve the same performance for a much smaller sample size (e.g. 75\% with 450 samples instead of 625) saving precious labelling time. As long as the forest has learnt enough to provide accurate measures of its prediction confidence (above training sets of 250 sample, which is 28\% of the entire data), all three active learning selective querying schemes (minimum margin, maximum entropy and maximum uncertainty) return better results than passive learning for most of training-test sizes. In the same fashion used for comparing overall models in subsection \ref{res-passive}, figure \ref{fig:figure-acc3} show the prediction result only for three SAT solvers. Interestingly, the higher the number of solvers, the greater the gains of active over passive learning (possibly due to the higher uncertainty). Figure \ref{fig:figure-acc10} is the specular image of figure \ref{fig:figure-acc3} for prediction over 10 solvers (the same used for the evaluations in table 3). When trying to predict the best solver out of 10 (figure \ref{fig:figure-acc10}), compared with the predictions over three solvers (figure \ref{fig:figure-acc3}), the accuracy gap between passive and active learning is much wider. For a training-test set of 300-589 samples (out of which 211 are selected in iteration using active learning) active learning achieves 12\% higher accuracy compared with passive learning, while with the same set sizes active learning achieves only 5\% higher accuracy when predicting the best out of three solvers. Also, it is interesting to notice that when predicting the best out of three solvers, the best performing active learning scheme is the maximum uncertainty approach (figure \ref{fig:figure-acc3}, green line). Differently, when increasing the portfolio size the best scheme is to the minimum margin approach (figure \ref{fig:figure-acc10}, light-blue line). This confirms that this approach works best when the solver portfolio is wider. Because it appeared to be more robust as the size of the portfolio solver increased, our recommended approach uses minimum margin as the selected active learning schema. However, when implementing the approach on a small number of solvers, the maximum uncertainty criteria should be considered as a valid alternative.

\section{Conclusions}\label{conclusions}
NP-Hard optimisation problems such as SAT are at the core of many applications in computer science. As research has shown that no single algorithm outperforms all the others, it is crucial to select the best algorithm for a given problem, which can be accurately done using supervised machine learning. However, creating a large dataset that can enable such learning systems is time-consuming and becomes harder the harder to problems to solve at hand.

Our contribution consists of using active learning to optimise the time needed to compile these time-expensive training sets. We trained an implementation of our approach using data from SAT competitions and show that we can build better or equally good algorithm selectors using less training data and saving precious solving time.

\section{Acknowledgement}\label{acknowledgement}

We acknowledge Dr Laszlo Vegh from the London School of Economics Department of Mathematics for his crucial supervision and advice of our work. We also acknowledge David Coller from the London School of Economics Master in Operational Research for contributing to the cooperation between the authors and the school.


\onecolumn 
\pagebreak
\section{Appendix}

\subsection{Appendix A}\label{appendix-A}
\begin{figure*}[ht]
  \centering
  \includegraphics[width=0.9\linewidth]{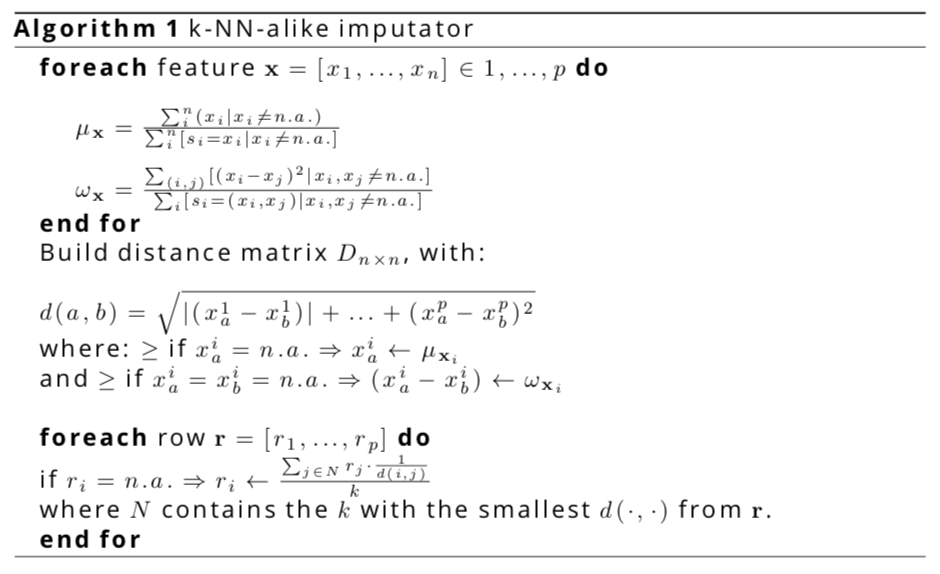}
  \caption{Algorithm A1. k-NN alike imputator}
  \label{fig:algorithmA1}
\end{figure*}

Figure \ref{fig:algorithmA1} shows the detailed implementation of the k-NN algorithm we developed to impute missing data in our framework.

\end{document}